\documentclass[runningheads]{llncs}
\usepackage[T1]{fontenc}
\usepackage{graphicx}
\usepackage[breaklinks,colorlinks]{hyperref}
\usepackage{color}

\usepackage{xspace}
\usepackage{bbding}  

\renewcommand{\arraystretch}{1.5}
\usepackage{tabularx}
\newcolumntype{Y}{>{\centering\arraybackslash}X}
\usepackage{multirow}

\usepackage{enumitem,amssymb}
\newlist{todolist}{itemize}{2}
\setlist[todolist]{label=$\square$}
\usepackage{pifont}
\usepackage{lipsum}

\makeatletter
\renewcommand{\paragraph}{%
  \@startsection{paragraph}{4}%
  {\z@}{0.8ex \@plus 1ex \@minus .2ex}{-1em}%
  {\normalfont\normalsize\bfseries}%
}
\makeatother

\usepackage{pgfplots} 
\pgfplotsset{compat=1.18}
\usepackage{diagbox}

\usepackage{tikz}
\usepackage{float}
\usetikzlibrary{positioning,calc}
\usetikzlibrary{angles, quotes}

\usepackage{subcaption}
\usepackage{amsmath}
\usepackage{booktabs}
\usepackage{url}

\usepackage{cleveref}
\usepackage{pgffor}

\newcommand{\mname}{pix2pockets\xspace}
\newcommand{\RL}{Shot Suggestion Model\xspace}
\newcommand{\CV}{Ball Location Model\xspace}

\newcommand{\mapn}{AP50\xspace}
\newcommand{\apn}{\mapn}

\newcommand{\imageinput}{I_{\text{in}}}
\newcommand{\imageoutput}{I_{\text{out}}}

\newcommand{\template}{T}
\newcommand{\homography}{H}

\newcommand{\dotsFound}{d}
\newcommand{\ballsFound}{b}
\newcommand{\postproc}{b'}

\newcommand{\approxCenter}{p}
\newcommand{\ballpos}{\widetilde{p}}

\newcommand{\state}{S}
\newcommand{\statespace}{\mathcal{S}}

\newcommand{\action}{A}
\newcommand{\actionspace}{\mathcal{A}}

\newcommand{\degree}{\alpha}
\newcommand{\force}{\rho}
\newcommand{\maskppo}{\text{PPO}_{\text{mask}}}

\newcommand{\oracleA}{$\oracle(\degree_{\text{rand}},\force_{\text{rand}})$}
\newcommand{\oracleB}{$\oracle(\degree_{\text{rand}},\force_{\text{max}})$}
\newcommand{\oracleC}{$\oracle(\degree_{\text{best}},\force_{\text{max}})$}
\newcommand{\maskA}{$\maskppo$}
\newcommand{\maskB}{$\maskppo(\force_{\text{max}})$}

\newcommand{\reward}{R}
\newcommand{\rewardspace}{\mathcal{R}}

\newcommand{\oracle}{\mathcal{O}}
\newcommand{\mirrortable}{\mathcal{M}}

\newcommand{\hitpoints}{HP}

\newcommand{\mypar}[1]{\vspace{0mm}\noindent\textbf{#1}}

\begin{document}

\title{\mname: Shot Suggestions in 8-Ball Pool from a Single Image in the Wild}
\titlerunning{\mname}
%


\author{\Envelope Jonas Myhre Schiøtt$^{1}$ \quad 
Viktor Sebastian Petersen$^1$ \quad 
Dim~P.~Papadopoulos$^{1,2}$ \quad \\
$^{1}$\,Technical University of Denmark
$^{2}$\,Pioneer Center for AI
\\
{\tt\small s204218@dtu.dk, s204225@dtu.dk, dimp@dtu.dk}
\\
\url{https://pix2pockets.compute.dtu.dk/}
}

%
\authorrunning{J. Schiøtt et al.}
%
%
\maketitle              
%

\begin{figure}[t]
    \centering
    \includegraphics[width=\linewidth]{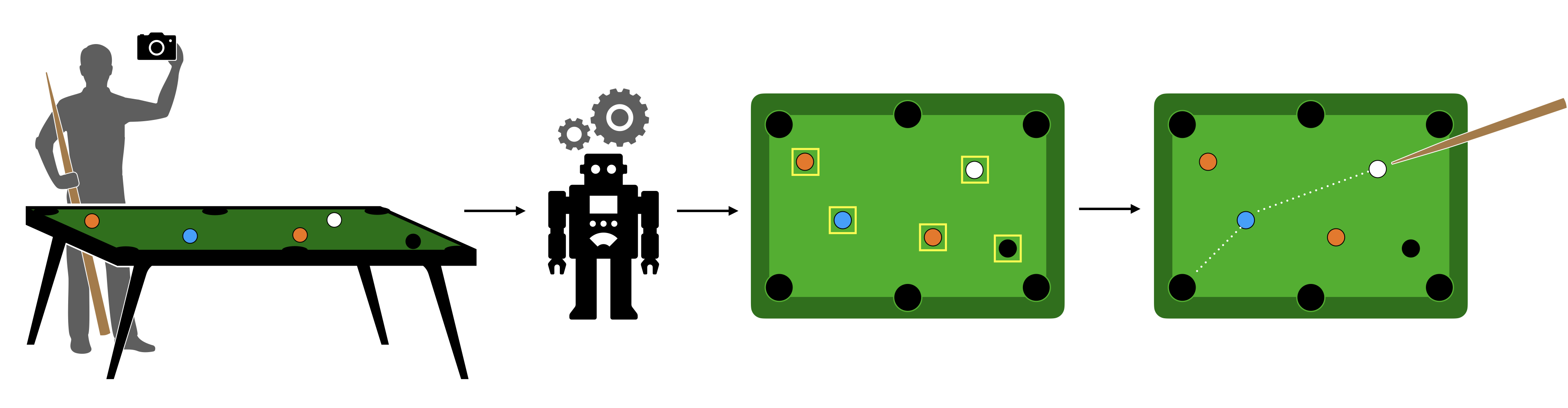}
    \caption{\textbf{\mname.} We introduce a new task for shot suggestions in pool games using a single input image. First, we detect the table and estimate the position of the balls.
    Then, they are fed into a pool environment, and a Reinforcement Learning agent predicts the best available shot (i.e., cue angle and shot power).}
    \label{fig:robot}
\end{figure}
\begin{abstract}

Computer vision models have seen increased usage in sports, and reinforcement learning (RL) is famous for beating humans in strategic games such as Chess and Go. In this paper, we are interested in building upon these advances and examining the game of classic 8-ball pool. We introduce \mname, a foundation for an RL-assisted pool coach. Given a single image of a pool table, we first aim to detect the table and the balls and then propose the optimal shot suggestion. For the first task, we build a dataset with 195 diverse images where we manually annotate all balls and table dots, leading to 5748 object segmentation masks.
For the second task, we build a standardized RL environment that allows easy development and benchmarking of any RL algorithm.
Our object detection model yields an \mapn of 91.2 while our ball location pipeline obtains an error of only 0.4 cm.
Furthermore, we compare standard RL algorithms to set a baseline for the shot suggestion task and we show that all of them fail to pocket all balls without making a foul move. 
We also present a simple baseline that achieves a per-shot success rate of 94.7\% and clears a full game in a single turn 30\% of the time.

\end{abstract}

\section{Introduction}
Artificial intelligence and reinforcement learning (RL) have proven to excel in complex games, such as Chess~\cite{AlphaZero}, Go~\cite{AlphaGo}, Starcraft~\cite{RL_Starcraft}, and Minecraft~\cite{minecraft}. 
Apart from board and video games, computer vision models have recently started playing an important role in sports with several applications in generating sports analytics~\cite{huang2006semantic} and analyzing game strategies and tactics~\cite{tuyls2021game,ramanathan2016detecting}.

While RL in sports is not as widespread~\cite{sun2020cracking,tuyls2021game}, table-based sports such as pool are a natural field for RL.
The 8-ball pool variant is a popular game played worldwide by millions of people. According to Guinness records, the most downloaded mobile game is the 8-Ball pool game with 800 Million downloads~\footnote{\url{https://tinyurl.com/ystrrdu9}}.
%
Therefore, obtaining an easy-to-use framework for shot suggestions would be a helpful tool for both amateurs and professionals. Although the research on RL-based pool agents is limited, many approaches for the task exist~\cite{Pool_Search,Pool_MonteCarlo,Pool_optimization}.

Several papers have proposed to detect the balls and edges on the pool table using a restricted setup~\cite{Pool_assist,Pool_ball_detection,SnookerReconstruction,8_ball_multi_object}. All these approaches try to solve two separate tasks; first locate the balls on the table and then suggest a good shot.
However, most of the approaches are limited to a single Bird's Eye View (BEV) of the table because their methods rely on a setup camera and traditional image processing techniques. This limits the training of such a model to a constrained situation, which is hard to obtain without any special equipment. Therefore, they fail to utilize modern deep learning models, which have shown great success in  sports~\cite{Football_ObjectDetection,Basketball_ObjectDetection,ballAndPlayer}. Consequently, to the best of our knowledge, an available dataset containing pool table images from various angles has not yet been established, which is needed for a model to work with images in the wild.

In this paper, we propose \textbf{\mname}, a foundation for an RL-assisted pool coach. Given a single image in the wild, we analyze the situation on the table and suggest a good shot for pocketing the next ball. For the first task, we build an image dataset containing different angles and views of the table, allowing a trained model to analyze any user image.
Our dataset contains 5748 manually annotated instances, with object bounding boxes for all balls and white dots~\footnote{\label{diamond-system}\url{https://www.libertygames.co.uk/pool-diamond-system/}} of the table.
For the second task, we establish a standardized RL environment compatible with the widely used Gymnasium framework~\cite{Gym}, which allows the easy training of any RL agent and the use of custom reward functions.

Our experimental results show that our detection model obtains a \mapn of 91.2\%. Using these detections, we build a method to find the accurate locations of the balls and map them into our RL environment. We obtain a mean location estimation error of 0.4 cm corresponding to only 7\% of the a pool ball diameter.
Furthermore, we experiment with standard RL algorithms~\cite{PPO,DDPG,TD3,A2C,SAC} on our developed RL environment.
Even though they succeed in easy situations where only two pool balls are present, they fail to suggest a successful shot when all balls are present.
We also present a simple algorithmic baseline that achieves a per-shot success rate of 94.7\% and clears a full table in a single turn 30\% of the time. 
%
We hope the release of the dataset and the environment will boost more work in the community for the game of pool.

\section{Related Work}
\label{sec:RelatedWork}

\mypar{Computer vision for sports.} Object detection systems are widely used in many sports such as football~\cite{Football_ObjectDetection}, basketball~\cite{Basketball_ObjectDetection} and handball~\cite{ballAndPlayer}, for generating sports analytics~\cite{huang2006semantic}, analyzing strategies~\cite{ramanathan2016detecting}, understanding broadcasts~\cite{giancola2018soccernet}, and forecasting future actions~\cite{felsen2017will}.
%
For the game of pool, there are several approaches to detect balls and edges on a table~\cite{Pool_ball_detection,SnookerReconstruction,Pool_assist,8_ball_multi_object,PoolRobot,PoolCueGuide,pool_smart_glasses}.
However, they are limited to a single BEV of the
table as they rely on a setup camera. This setup is hard to use in practice with images from various angles and lighting conditions. 
Moreover, these approaches use classic image processing based on thresholding~\cite{PoolCueGuide}, Hough transformation~\cite{PoolRobot,SnookerReconstruction}, and morphological operators~\cite{SnookerReconstruction}.
Instead, we are interested in detecting pool tables and balls on images in the wild with diverse angles and lighting conditions.




\mypar{Learning to play board games.}
In 1996, the chess match between Garry Kasparov and the IBM computer DeepBlue was a milestone. Even this brute-force approach showed that artificial intelligence can catch up to human intelligence and defeat intellectual champions.
%
Nowadays, most approaches rely on powerful RL models. One of the first examples is the 1992 IBM TDgammon~\cite{TD-gammon}, which used TD-lambda for playing backgammon. Recently, AlphaGo~\cite{AlphaGo} and AlphaZero~\cite{AlphaZero} have become increasingly advanced, surpassing human performance.

\mypar{Learning to play video games.}
Other RL approaches focus on video games ranging from simple Atari games~\cite{atari} like Pong or Space Invaders to complicated modern games like Counter-Strike~\cite{RL_counterstrike}, Minecraft~\cite{RL_Minecraft}, and Starcraft II~\cite{RL_Starcraft}. An example of RL in video games is studied in~\cite{atari}, in which only the raw 
pixels from seven classic Atari games are used as input to an RL agent. This study shows that with most games, state-of-the-art methods could be implemented to achieve 
performance better or comparable to an expert human player~\cite{atari}. Other papers~\cite{RL_counterstrike,RL_FPS} build RL agents that perform well in FPS games. One uses behavioral cloning from a large dataset consisting of Counter-Strike videos~\cite{RL_counterstrike}, while the other uses a modified Q-learning algorithm named RETALIATE~\cite{RL_FPS}.

\mypar{Learning to play pool.}
Pool also requires strategy and outcome predictions. One attempt to create an agent for a pool environment is made in a series of YouTube videos~\footnote{\url{https://github.com/packetsss/youtube-projects/tree/main/pool-game}}. However, this mainly showcases results, and the training implementation is limited.
Another example is a playable pool implementation in pygame~\footnote{\url{https://github.com/russs123/pool\_tutorial}}.
Other simulations in online games have the player compete against an AI, which often uses a set of predefined strategies, vector calculations, or search trees~\cite{Pool_MonteCarlo,Pool_Search,Pool_optimization}.
Instead, we implement a standardized RL environment which handles a variety of predefined agents using the Gymnasium library~\cite{Gym}.

\renewcommand{\arraystretch}{2}

\begin{figure*}[t]
  \centering
  \begin{subfigure}{0.38\linewidth}
    \raggedright
    \input{sec/tikz/datasetex}
    \caption{\textbf{Dataset example images}}
    \label{fig:dataset}
  \end{subfigure}
  \hspace{-1em}
  \begin{subfigure}{0.3\linewidth}




\begin{tikzpicture}
    \begin{axis}[
        width=\linewidth,
        height=5.2cm,
        clip=false,
        separate axis lines,
        axis on top,
        ytick={1,2,3,4,5},
        y tick style={draw=none},
        x tick style={draw=none},
        yticklabels={Dot,Black,Cue,Solid,Stripes},
        xmin=0,
        xmax=6500,
        xbar,
        major x tick style = transparent,
        xticklabel=\empty,
        nodes near coords,   
        nodes near coords style={anchor=west,#1,black},
        every axis plot/.append style={
          bar width=14pt,
          bar shift=0pt,
          fill
        }
      ]
      \addplot[blue]coordinates {(891,5)};
      \addplot[green]coordinates{(976,4)};
      \addplot[red]coordinates{(194,3)};
      \addplot[black]coordinates{(191,2)};
      \addplot[magenta]coordinates{(3496,1)};


    \end{axis}
  \end{tikzpicture}
\vspace{-0.8em}
    \vspace{-1em}
    \caption{\textbf{Dataset statistics}}
    \label{fig:statistics}
  \end{subfigure}
  \hfill
  \begin{subfigure}{0.308\linewidth}
    \centering
    \includegraphics[width=\linewidth]{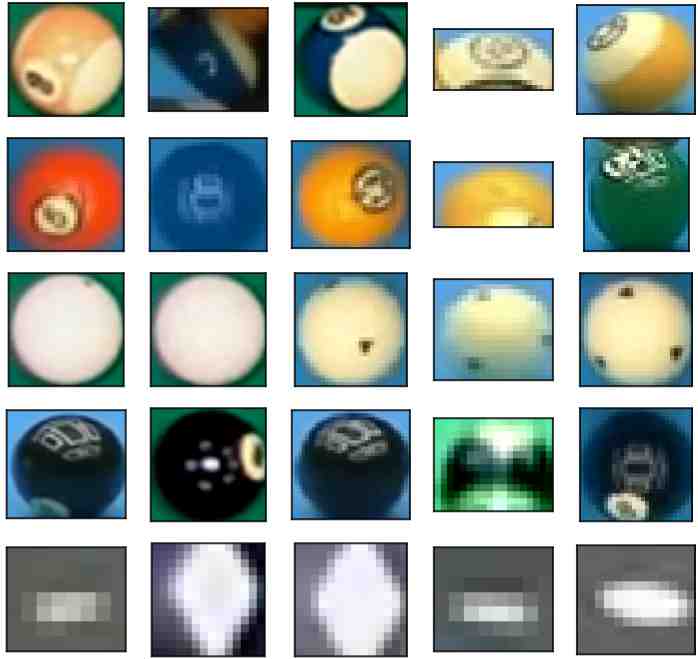}
    \caption{\textbf{Object examples}}
    \label{fig:objects}
  \end{subfigure}
  \caption{\textbf{Our Dataset.}
  (a) It contains 195 annotated images of tables captured from various angles with diverse lighting conditions. 
  (b) We annotate 5748 objects with accurate segmentation masks. The maximum number of objects in the image varies from class to class.
  (c) Bounding box annotated examples. Note how sometimes the balls are not completely visible from the given view.}
  \label{fig:data+stats}
\end{figure*}

\renewcommand{\arraystretch}{1.5}

\section{Dataset}
\label{sec:Dataset}
We present our dataset collected to train object detection models in a pool game. The images are obtained from various online videos from different 8-ball championships. Most of these frames are taken from the \textit{Predator World 8-Ball Championship}~\footnote{\url{https://www.youtube.com/@ProBilliardTV}}, as these are high-quality videos of many different angles filmed using BEV cameras and camera jibs. This allows us to gather diverse images with different views of the table. Example images are shown in Fig.~\ref{fig:dataset}.
We manually annotate the bounding boxes of all balls (cue, black, striped, and solid) on every image. To detect the table, we annotate the bounding boxes of white dots around the table. We found that this is a better choice than annotating the whole surface of the table, as the dots are located identically on all tables according to the \textit{Diamond System}\textsuperscript{\ref{diamond-system}} and therefore uniquely identify the boundaries of the table. All annotations are obtained in Roboflow Annotate.
%
Our dataset consists of two sets. The main dataset consists of 195 images (5748 bounding box annotations) used to train the detection model. The additional dataset contains 52 images (1624 annotations) of pool situations, where multiple images show the same table from different angles. There are 25 pool situations in total, which we use to study the projection error from view to view (Sec.~\ref{secCVexp}).
Examples of the annotated bounding boxes (balls and dots) are shown in Fig.~\ref{fig:objects}, while detailed dataset statistics are shown in Fig.~\ref{fig:statistics}. Note that not all images have the cue ball, 8-ball or all dots, as they can be hidden behind the players in the images.

\begin{figure*}[t]
    \centering
    \input{sec/tikz/pipeline}
    \caption{\textbf{Full pipeline.} 
    The input image $\imageinput$ is run through the \CV to estimate the ball positions on the table, which is then handed to the \RL.
    First, we obtain the dot detections $\dotsFound$ and the ball detections $\ballsFound$ on $\imageinput$. We use $\dotsFound$ to find the table lines and thus estimate a mapping $\homography$ from $\imageinput$ to a template $\template$. Then, we use $\homography$ to estimate the center point $\approxCenter$ for the balls $\ballsFound$, resulting in the positions $\ballpos$ for the environment.
    The \RL sends the state $\state \in \statespace$ to the agent, suggesting the $\action \in \actionspace$. During training, the environment evaluates the action, and the agent receives a reward $\reward \in \rewardspace$.}
    \label{pipline}
\end{figure*}

\section{Method}
\label{sec:Methods}

We propose \mname, a method for generating shot suggestions given a single input image $\imageinput$. The overall pipeline is depicted in Fig.~\ref{pipline}. Our model can be split into two components: the \CV, and the \RL.

\subsection{\CV}

The \CV provides precise ball locations given $\imageinput$. 
We use a pre-defined template, $\template$, that shows a BEV of a pool table. The template $\template$ matches the dimensions in an RL environment, and the goal is to map the ball centers in $\imageinput$ to $\template$ using an estimated homography matrix $\homography$. First, we use an object detection model to obtain the ball and dot detections on the table. Then, we perform a line estimation approach on the dots to ensure the table is rotated correctly and estimate $\homography$. We also use $\homography$ to estimate the ball centers.
Finally, we map the ball centers to $\template$, and use them at the \RL.


\mypar{Dot detections.}
To locate the table in the image, we obtain a set of dot detections $\dotsFound$. Pool tables follow the diamond system~\textsuperscript{\ref{diamond-system}}, which ensures that the dots will be in a special format, allowing us to learn the dimensions of the real table. 

\mypar{Line estimation.}
We use a linear line interpolation from $\dotsFound$, to identify the four sides of the table, thereby dividing $\dotsFound$ into four subsets. We iteratively use Ransac~\cite{Ransac} to find the line that interpolates the most points. We remove the inliers from each iteration and end up with the four lines (four table sides). To ensure that the line estimation is correct, we find all the intersections between them and make sure that only four intersection points are within the table. Furthermore, we find the intersections of these lines and add them to the set of points $\dotsFound$. 
To estimate $\homography$, the order of the elements in $\dotsFound$ needs to match the order in $\template$. This is achieved by sorting the four sets of points (one for each line) by their mean $(x,y)$ values and then sorting the individual points within each set by $x$.
After this, we check whether the first line contains 3 (short side) or 6 (long side) points since this indicates if we need to permute the ordering to ensure that the narrow table side is not mapped to the long side and vice versa. 


\mypar{Homography.}
Given the template $\template$ and found dots $\dotsFound$, we estimate a mapping from $\imageinput$ to $\template$ with $\homography$. In theory, we only need 4 points to estimate $\homography$, but we use all 22 available points to make the estimation more robust.

\mypar{Ball detections.}
In parallel, we obtain a set of ball detections $\ballsFound$ from $\imageinput$. They are represented as bounding boxes $(x,y,w,h,c)$, where $(x,y)$ is the upper-left corner of the box, $(w,h)$ is the width and height, and $c$ is the predicted label.

\mypar{Approximate ball center points.}
To map the balls $\ballsFound$ to $\template$, we determine a single point $\approxCenter_i$ for each $\ballsFound_i$. 
This point should be in the center of the bounding box for BEV images, whereas for angled images, lie closer to the top of the ball. 
To approximate $\approxCenter$, we approximate the camera angle using $\homography$ and map it linearly between the center point and the topmost point of the bounding box.



\mypar{Ball positions.}
To correctly position the balls inside the environment we use $\homography$ to map each $\approxCenter_i$ to a point $\ballpos_i \in \template$ used as ball positions for the balls in Sec.~\ref{sec:5:rl}.

\subsection{\RL} \label{sec:5:rl}

The \RL provides shot suggestions based on the given ball positions. We utilize RL to achieve this. For the RL agent to give shot suggestions, we create an environment in which it can gain experience.

\mypar{Environment.}
The environment is built using the widely used Python module \verb'Gymnasium'~\cite{Gym}. Here, we can define the layout and rules for which the RL agent can be trained. Our environment is created to be identical to a real pool table. We ensure this by using the dimensions of all elements from the top-view images from our dataset, such that the environment follows the actual pool table dimensions.
The same applies to the size of the pockets relative to the balls and their locations.
Due to the complexity of this task, we choose to simplify the physics as much as possible. We assume that the cue ball is always hit in its center and that every collision is perfectly elastic. We use the Python module \verb'Pymunk' to simulate the physics. 
Since the environment is fully observable, the state $\state \in \statespace$ contains the whole observation. That is $\state_t = (x_1,y_1,c_1, ... x_{16}, y_{16}, c_{16})_t$, i.e., the position and class of all the balls at timestep $t$. The state space $\statespace$ contains all possible pixel locations inside the table boundaries for all balls, with the conditions that (a) the center of the ball cannot be closer to a cushion than its radius, and (b) any two balls cannot be closer to each other than their radii combined. To visualize and run the environment, we use the Python module \verb'Pygame'.


\mypar{RL agent.}
Given $\state_t$ at timestep $t$, the agent suggests a single shot as an action $\action_t$. The action space $\actionspace = \{\degree, \force\}$ where $\degree$ is the direction and $\force$ is the power of the shot.
During training, $\action_t$ is evaluated by
$\reward: \statespace \times \actionspace \rightarrow \rewardspace$ where $\rewardspace$ is the set of possible rewards from the reward function $\reward$ described in Sec.~\ref{secRLexp}.

\paragraph{Output.}
The model output is an image $\imageoutput$ showing the state $\state$ from the environment with a depiction of the suggested action $\action$.

\section{Experiments}
\label{sec:Experiments}
In this section, we present our experimental results. First we present experiments for the \CV, and then for the \RL.

\subsection{\CV} \label{secCVexp}
We use a pre-trained YOLOv5 model~\cite{YOLO}, which we finetune to our dataset. We split our dataset into 155 training, 20 validation, and 20 test images. The images are resized to $640 \times 640$, and we use a batch size of 20, a constant learning rate of 0.01, and 2000 epochs, as fewer epochs lead to a model that can't distinguish the cue from the 8-ball. To evaluate the detection model, we use precision, recall, F1-score, and the average precision with an IoU-threshold of 50 (\mapn). 

\newcommand{\titx}[1]{\small#1}
\newcommand{\al}{\phantom{1}}

\begin{table}[t]
    \centering
    \caption{\textbf{Ball and table detection.} The performance of the object detection model before and after post-processing (before $\vert$ after). The post-processing step increases the precision by about 10\%.
    }
    
    {\small
    \begin{tabularx}{\linewidth}{lYYYYYY}
    \toprule
        \titx{$b\ \vert \  b'$} & \titx{Stripes} & \titx{Solids} & \titx{Cue} &  \titx{Black} & \titx{Dots} & \titx{Average} \\
    \midrule 
        \titx{Precision} & $84.0\ \vert \ 97.8$ & $90.0\ \vert \ 95.3$ & $86.4\ \vert \ 95.0$ & $85.7\ \vert \ 100$ & $83.5\ \vert \ 90.8$ & $85.9\ \vert \ 95.8$ \\
        \titx{Recall}    & $89.9\ \vert \ 89.9$ & $89.0\ \vert \ 89.0$ & $95.0\ \vert \ 95.0$ & $94.7\ \vert \ 94.7$ & $91.6\ \vert \ 90.8$ & $92.0\ \vert \ 91.9$ \\
        \titx{F1}        & $86.8\ \vert \ 93.7$ & $89.5\ \vert \ 92.0$ & $90.5\ \vert \ 95.0$ & $90.0\ \vert \ 97.3$ & $87.4\ \vert \ 90.8$ & $88.8\ \vert \ 93.8$ \\
        \titx{\apn}      & $90.0\ \vert \ 91.4$ & $89.9\ \vert \ 90.4$ & $92.2\ \vert \ 92.3$ & $92.2\ \vert \ 92.4$ & $89.5\ \vert \ 89.3$ & $90.8\ \vert \ 91.2$ \\
        \titx{AP50:95}   & $79.1\ \vert \ 80.4$ & $80.7\ \vert \ 80.9$ & $83.6\ \vert \ 84.0$ & $84.9\ \vert \ 85.4$ & $51.0\ \vert \ 51.4 $ & $75.9\ \vert \ 76.4$\\
    \bottomrule
    \end{tabularx}
    }
    \label{classmAP}
    \vspace{-5mm}
\end{table}

\mypar{Detection Results.}
The object detection results are shown in Tab.~\ref{classmAP}. We observe qualitatively that many of these errors are caused by detections that are not possible (e.g., detecting more than 18 dots or more than one cue). In other cases, balls are detected as multiple classes, resulting in a very high overlap of several objects. Since we know this can never happen, we constrain the model by post-processing the detections.
We establish a post-processing procedure $\ballsFound \rightarrow \postproc$, that enhances the detection results. The post-processing first runs a class agnostic non-maximum-suppression (NMS) to eliminate detections with high overlap. Then, the highest confidence detections are kept while respecting the total number of class instances in an image (i.e., 18 balls). 
We see that the use of post-processing generally improves performance for all metrics. Due to the removal of detections, the post-processing sometimes removes correct but low-confidence detections, resulting in a lower recall, but with a precision increase of 10 percentage points, we obtain a more robust model.

\mypar{Hough Circles.} We compare our model with a traditional image-processing method (Hough Circles). We first apply an adaptive threshold on the image and then use the 
Hough Circles approach
with $dp=1$, $minDist=10$, $param1=300$, $param2=0.7$, $maxRadius=30$. The $minRadius$ is set to 0 for dot detections and 15 for balls. After detecting the circles, we perform an NMS using the radius as a confidence score. The Hough Circles achieve an AP50:95 of 0.26 without dots and 0.21 with dots, while our model, when disregarding classes, achieves 0.82 without dots and 0.67 with dots. In addition to being a better detector, our model can also classify them, which the Hough Circles is not able to.

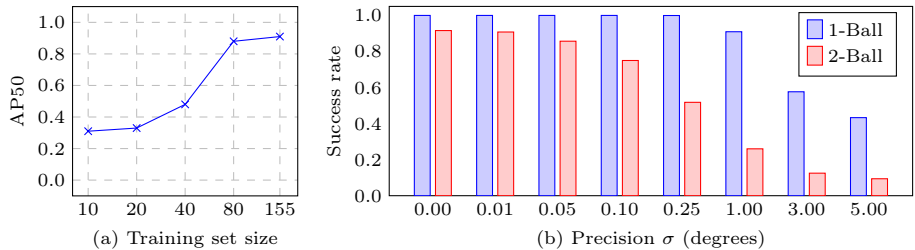
\begin{figure}[t]
    \centering
    \begin{minipage}[t]{.30\linewidth}
\raggedright
\begin{tikzpicture}
\scriptsize
\begin{axis}[
    width=1.25\linewidth,
    height=4.1 cm,
    xlabel={(a) Training set size}, xlabel near ticks, 
    ylabel={\mapn}, ylabel near ticks,
    xmin=8, xmax=200,
    ymin=-0.1, ymax=1.1,
    x tick style={draw=none},
    y tick style={draw=none},
    xtick={10,20,40,80,155},
    ytick={0.0,0.2,0.4,0.6,0.8,1.0},
    xticklabels={10,20,40,80,155},
    yticklabels={0.0,0.2,0.4,0.6,0.8,1.0},
    ymajorgrids=true,
    xmajorgrids=true,
    grid style=dashed,
    xmode=log,
]
\addplot[color=blue,mark=x]
    coordinates {
    (10,0.31)
    (20,0.33)
    (40,0.48)
    (80,0.88)
    (155,0.91)
    };
\end{axis}
\end{tikzpicture}
\end{minipage}
\hspace{-1em}
\hfill
\begin{minipage}[t]{.70\linewidth}
\raggedleft
\begin{tikzpicture}
\scriptsize
    \begin{axis}[
        width=1\linewidth,
        height=4.1cm,
        ymax=1.05,
        ymin=0.0,
        ybar,
        xlabel={(b) Precision $\sigma$ (degrees)}, 
        xlabel near ticks,
        ylabel={Success rate},
        ylabel near ticks,
        legend pos=north east,
        legend image code/.code={\draw [#1] (0cm,-0.1cm) rectangle (0.2cm,0.15cm); },
        symbolic x coords={A,B,C,D,E,F,G,H}, 
        xticklabels={0.00,0.01,0.05,0.10,0.25,1.00,3.00,5.00},
        ytick={0.0,0.2,0.4,0.6,0.8,1.0},
        yticklabels={0.0,0.2,0.4,0.6,0.8,1.0},
        xtick=data,
        x tick style={draw=none},
        y tick style={draw=none},
        xticklabel style={yshift=4.2pt},
        bar width=6pt,
        ]
                \addplot+[ybar, fill=blue!20] plot coordinates {
                (A, 1.0) (B, 1.0) 
                (C, 1.0) (D, 1.0) 
                (E, 0.9995) (F, 0.9095) 
                (G, 0.5765) (H, 0.433)
                }; 
                \addplot+[ybar, fill=red!20] plot coordinates {
                (A, 0.916) (B, 0.908) 
                (C, 0.857) (D, 0.750) 
                (E, 0.518) (F, 0.260) 
                (G, 0.125) (H, 0.094)
                };
        \addlegendentry{1-Ball}
        \addlegendentry{2-Ball}
    \end{axis}
\end{tikzpicture}
\end{minipage}
    \caption{\textbf{(a) Training size.} The \mapn of models with different training set sizes, showing diminishing gains after 80 images. 
    \textbf{(b) Shot Accuracy.} To determine the shot precision, we test the performance for different $\sigma$ values. In the 1-Ball environment, a precision of 0.25 degrees is enough to pocket the ball. Ball-ball interaction requires larger precision (0.01 degrees) when using additional balls. }
    \label{fig:datasize}
\end{figure}

\mypar{Training size.}
To test the training performance using the dataset, we perform ablation on the size of the training set by training several YOLOv5 models with varying training set sizes, shown in Fig.~\ref{fig:datasize}a. We observe that \mapn is very low ($<$50\%) when training with less than 50 images. Using 80 images yields much better results close to our full model, which yields an \mapn of 91\%.

\mypar{Table size.}
The further away one is from the table, the harder it is to distinguish the ball classes. We evaluate this by calculating the \mapn of each test image as a function of the table area in the image. The \mapn is above 95\% when the table covers at least 40\% of the image. When the table occupies a smaller area such that objects occupy less than 8$\times$8 pixels, recognizing their categories becomes challenging, and \mapn drops significantly to about 50\% in these few cases.

\begin{figure*}[t]
\captionsetup[subfigure]{labelformat=empty}
\centering
\begin{subfigure}{0.46\linewidth}
\centering
\input{sec/tikz/ballzoom1}
    \caption{\textbf{Best case.} Mean table shift: 0.22 cm.}
    \label{fig:shift_front}
\end{subfigure}
\hfill
\begin{subfigure}{0.46\linewidth}
\centering
\input{sec/tikz/ballzoom2}
    \caption{\textbf{Worst case.} Mean table shift: 0.76 cm.}
    \label{fig:shift_45}
\end{subfigure}
\caption{\textbf{Projection error.} To estimate the projection error, the front-view and 45-view projections are compared to the top-view ground truth. The projection result is shown on the top-view image for accuracy assessment. The blue lines indicate the distance from the estimated center point $\approxCenter$ to the ground truth. The mean shift is compared to the table length in $\template$, and scaled to a regular 9ft table.}
\label{fig:shift}
\end{figure*}

\mypar{Projection Error.} \label{sec:projection-error}
To evaluate $\homography$, we establish a controlled experiment, where we use two different frames for the same pool situation: one BEV and another view using the additional dataset of 25 sets of situations from multiple angles. Here, we treat the BEV image as ground truth, as the template transformation is trivial. Then, we compare the results of the ball locations from the two estimated homographies, and the shift error is calculated as the distance from the ball centers using the BEV homography to the centers using the other homography.
We obtain a mean error of 0.4 cm (good and bad examples in Fig.~\ref{fig:shift}). We observe that the error is relatively small compared to the size of a real ball (5.7 cm).

\subsection{\RL}\label{secRLexp}

In this section, we discuss the specific choices taken in building the environment and then we present the results and evaluations of the trained RL models.

\mypar{Environment settings.}
We set the screen size to 410 and 735 pixels. These lengths consist of the actual table itself, but also the width of the edges. The ball's radius is set to 7 pixels, and the pocket's radius to 15 pixels. Deducting the sides of the table, we have a table state space with $\vert\statespace\vert$ $=$ $337\times662$
valid ball positions. When training, we generate states at random so that there is no overlap with neither another ball nor a cushion. As precision lower than 0.01 degrees has proven to result in lower performance, we use a multidiscrete action space with 36.000 equidistant angles and 30 values for the shot power matching the power of a shot in a real tournament, leading to an action space of size
$\vert\actionspace\vert = \vert\degree\vert \times \vert\force\vert = 36000\times30$. The target balls will be denoted "blue" balls.

\mypar{Environments.}
We experiment with 3 environments: 1-ball, 2-ball, and All-ball.
In 1-ball, only the cue ball is present, and the goal is to pocket it. 
In 2-ball, the cue and 8-ball are present, and the goal is to hit the cue ball and pocket the 8-ball.
All-ball simulates a full game, and the goal is to first pocket all blue balls, and then the 8-ball, without mistakes. In all cases, the environment is reset if a shot is unsuccessful, whereas a new shot is awarded if a ball is pocketed.

\begin{figure*}[t]
    \centering
    \input{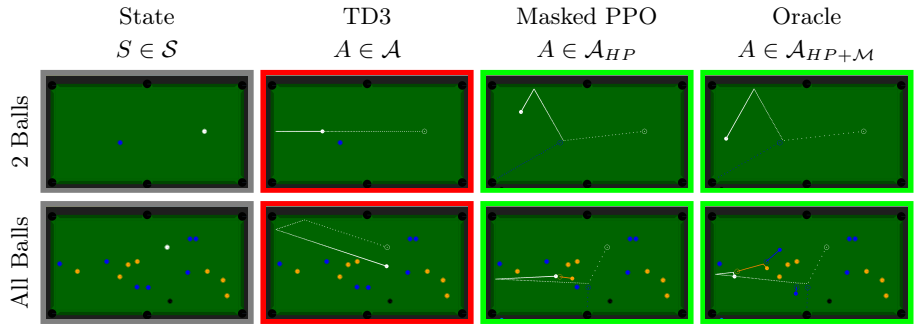}
    \caption{\textbf{Example of shot suggestions.} The first column shows two initial states. The second, third, and fourth columns show the action predictions of TD3, Masked PPO, and the Oracle, respectively. Each action image is highlighted, indicating whether the shot was successful (green) or not (red).}
    \label{fig:RLplot}
\end{figure*}

\mypar{Baseline models.}
We use standard RL algorithms provided in the SB3 library~\cite{stable-baselines3}. These include the Proximal Policy Optimization (PPO)~\cite{PPO}, the Deep Deterministic Policy Gradient (DDPG)~\cite{DDPG}, the Twin Delayed DDPG (TD3)~\cite{TD3}, the Advantage Actor Critic (A2C)~\cite{A2C}, and the Soft Actor-Critic (SAC)~\cite{SAC}. In addition, we use Masked PPO~\cite{MaskedPPO}, which masks out invalid actions from PPO. 

\mypar{Reward function.}
We reward the agent for winning the game by +100 and penalize it for losing it by -100. The agent must pocket all blue balls and then the 8-ball to win. To lose, the agent must pocket the cue ball or the 8-ball when any blue balls remain. To incentivize hitting the blue balls, the agent gets +10 for each blue ball hit and +50 for each blue ball pocketed. If a blue ball is hit, it gets a reward depending on the minimum angle between the hit balls' velocity and the pockets as described by $r(v) = 1000/(v+10)-50$, where $v$ is the angle mentioned above. If the agent doesn't hit anything, it is penalized depending on the distance to the closest blue ball when all balls lie still again, as described by $r(d) = -50 d/D$, where $d$ is the distance, and $D$ is the length of the diagonal of the table. If the agent pockets the cue ball, it is penalized by -80. Lastly, the reward is clipped to to the interval $[-210, 210]$ and normalized to $[-1, 1]$.

\mypar{Hitpoints.}
We introduce a calculation of all possible directions for a successful shot that pockets a target ball.
These directions are indicated by the point the cue ball needs to hit to deliver an impulse to the given ball in the correct direction. We refer to these points as hitpoints $\hitpoints$. Hitpoints can be used for action masking. By limiting the agent to shoot in directions that lead to correctly pocketed balls, we can significantly reduce the action space from the original $\vert\actionspace\vert = 36000 \times 30$ to the subspace $\vert\actionspace_{\hitpoints}\vert = \vert\hitpoints\vert \times 30$.

\mypar{Oracle.}
We propose our Oracle model $\oracle$ to quantize the value of the hitpoints and determine the best option by two factors. First, a change in direction due to an angled collision makes a shot harder, and thus, we consider the cosine similarity of the proposed direction vector for the cue ball and the ball to hit. Second, we use the target window (the angle from the ball to the two sides of the pocket hole) as this determines the precision needed to fulfill the shot. 
The Oracle can suggest the best shot directly without using RL or quantize the shots for the reward function to advance training further.

\newcommand{\arwlen}{.5}
\newcommand{\tabl}{1.1}

\begin{figure}[t]
    \centering
\tikzstyle{arrow} = [thick,->,>=stealth]
\begin{tikzpicture}
 \node (im) [] {\includegraphics[width=.7\linewidth]{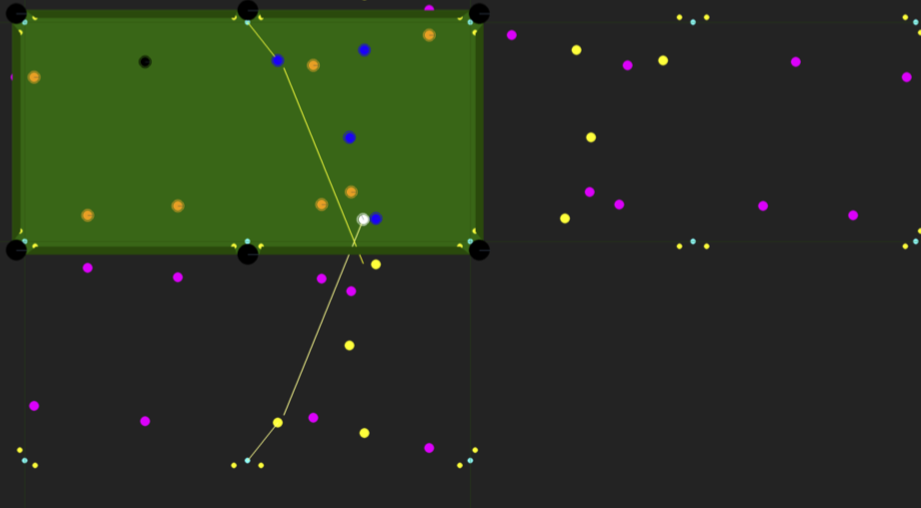}};
\draw [arrow] ($(im.north)+(-2.25,0)$)-- ($(im.north)+(-2.25,\arwlen)$);
\draw [arrow] ($(im.west)+(0,\tabl)$) -- ($(im.west)+(-\arwlen,\tabl)$);
\end{tikzpicture}
\caption{\textbf{Mirror table.} To include kick and bank shots, we establish an approach to mirror the balls and pockets to a separate space on each table side. Aiming for a ball or pocket in the mirrored space $\mirrortable$ is equivalent to hitting a cushion and then a ball on the real table. The cyan points indicate the aiming points.
}
    \label{fig:mirror-crop}
\end{figure}

\mypar{Introduce Mirror table.}
While direct shots are typically easier,
kick or bank shots (i.e., hitting the cushions during the shot) can offer better options.
To include these options, we expand the hitpoint space $\statespace_{\hitpoints}$ by introducing mirrored table versions $\mirrortable$ on each side (Fig.~\ref{fig:mirror-crop}).
%
By mirroring the table, the cue ball direction can be pointed towards a ball in $\mirrortable$ for kick shots, resulting in hitting the cushion and then going to the desired ball in $\statespace$. Similarly, bank shots can be simulated by aiming at balls pocketing in $\mirrortable$. 
This expands the hitpoint set and improves the chances of finding a good shot.
We also apply a penalty multiplier (0.33) for each cushion hit, as direct shots are easier to perform in reality. 
%

\mypar{Results.}
We train each baseline model at each environment for 500.000 timesteps using three-layer linear policy networks.
We add action noise for algorithms with continuous action space (TD3, SAC, and DDPG) to help with exploration. All other hyperparameters are the default ones from SB3. In Tab.~\ref{tab:env-results}, we list the completion percentages for every model in each environment. A large gap between the 1-ball and the 2-ball environment indicates a significant difficulty spike between the two. No standard RL model manages to complete the all-ball environment, although they pocket at least one ball 4-13\% of the time. Masked PPO outperforms the other baselines, but only Oracle completes the all-ball environment with a 30\% success rate. 
As a reference, a professional 8-ball player can ``break and run'' 39.2\% of the time~\cite{break_and_run}.
In Fig.~\ref{fig:RLplot}, we observe that Masked PPO and Oracle succeed in contrast to TD3.
This may be due to the large action space lowered for Masked PPO and completely removed for Oracle. 
The simple reward system may also be a reason for the low performance of the RL agents. 

\newcommand{\dshow}{\%}

\newcommand{\envA}{1-ball}
\newcommand{\envB}{2-ball}
\newcommand{\envC}{all-ball}
\newcommand{\envD}{Full Turn}

\begin{table}[t]
{\scriptsize
    \centering
    \caption{\textbf{Success-rate using various RL algorithms.} 
    We train all baselines for one shot in the three environments, with little success in the latter two.} 
    \begin{tabularx}{\linewidth}{lYYYYYY}
    \toprule
    Env & Random & PPO & TD3 & A2C & DDPG & SAC \\
    \midrule
    \envA  & 23.6\dshow & 84.6\dshow & 97.4\dshow & 35.0\dshow & 99.8\dshow & 88.6\dshow \\
    \envB  & 0.90\dshow & 1.10\dshow & 6.30\dshow & 0.00\dshow & 5.50\dshow & 9.90\dshow \\
    \envC & 4.30\dshow & 12.8\dshow & 7.60\dshow & 10.6\dshow & 6.40\dshow & 5.90\dshow \\
    \bottomrule
    \end{tabularx}
    \label{tab:env-results}
    }
\end{table}

\begin{table}[t]
\vspace{-1.5em}
{\scriptsize
    \centering
    \caption{\textbf{Success-rate using hitpoints.} Using the same environments, we use hitpoints (direct $\vert$ direct,mirror) as action masks. We also test stages of using the oracle with randomness or evaluated maxima. In addition, we test the completion of a full turn (all blue balls and then the black ball).} 
    \begin{tabularx}{\linewidth}{lYYYYY}
    \toprule
    Env & \maskA & \maskB & \oracleA & \oracleB & \oracleC \\
    \midrule
    \envA  & $100\dshow\ \vert \ 100\dshow$ & $100\dshow\ \vert \ 100\dshow$ & $84.3\dshow\ \vert \ 86.2\dshow$ & $100\dshow\ \vert \ 100\dshow$ & $100\dshow\ \vert \ 100\dshow$ \\
    \envB  & $63.2\dshow\ \vert \ 68.5\dshow$ & $66.4\dshow\ \vert \ 67.5\dshow$ & $65.9\dshow\ \vert \ 34.4\dshow$ & $88.5\dshow\ \vert \ 59.9\dshow$ & $95.2\dshow\ \vert \ 98.7\dshow$ \\
    \envC  & $49.7\dshow\ \vert \ 60.7\dshow$ & $53.4\dshow\ \vert \ 57.1\dshow$ & $68.7\dshow\ \vert \ 37.1\dshow$ & $84.6\dshow\ \vert \ 57.3\dshow$ & $90.2\dshow\ \vert \ 94.7\dshow$ \\
    \midrule
    \envD  & $0.00\dshow\ \vert \ 0.00\dshow$ & $0.00\dshow\ \vert \ 0.00\dshow$ & $0.00\dshow\ \vert \ 0.00\dshow$ & $1.50\dshow\ \vert \ 0.30\dshow$ & $13.8\dshow\ \vert \ 30.0\dshow$ \\
    \bottomrule
    \end{tabularx}
    \label{tab:env-results2}
    }
\end{table}

\mypar{Ball shift effect.} To measure the impact of the projection error on a good shot, we set up a 2-ball environment and make the Oracle find a shot. Then, we shift the target ball some distance in a random direction and let the shot play out. The distances shifted are sampled uniformly either from the mean shifts found in Sec.\ref{sec:projection-error} (shift-1) or a fixed distance of 2.5 cm (shift-2). 
We find that the Oracle using direct hitpoints drops from 97.06\% success to 77.88\% when the ball has been shifted by shift-1. However, when including the mirrored hitpoints, the drop is from 98.46\% to 84.52\%. For shift 2, the new percentages are 31.38\% and 33.78\%, respectively. This shows that the small projection error of our \CV does not change the results drastically, but a larger one will.
%
    

\newpage

\section{Conclusion}
\label{sec:Conclusion}
We presented a foundation for an RL-assisted Pool Coach named \mname. Given an image, the model can suggest a shot by predicting the best direction and power. We built a dataset to enable the training of a pool-ball detector such that it can detect balls in any user image captured in the wild.
We created simplified and open-source RL environments that provide standard benchmarks for training any RL algorithm. 
We show that pocketing the cue ball directly is easy, but 
pocketing all balls without making a foul move is much harder to accomplish.
Our work marks a new direction for systematically benchmarking algorithms to solve the game of pool and assist in training amateur and professional players.

\noindent\textbf{Acknowledgements.} D. Papadopoulos was supported by the DFF Sapere Aude Starting Grant "ACHILLES".



%
%
%

    \clearpage
    {\small
    \bibliographystyle{splncs04}
    \bibliography{main}}

\end{document}